\documentclass[lettersize,journal]{IEEEtran}
\usepackage{amsmath,amsfonts}
\usepackage{algorithmic}
\usepackage{algorithm}
\usepackage{array}
\usepackage[caption=false,font=normalsize,labelfont=sf,textfont=sf]{subfig}
\usepackage{textcomp}
\usepackage{stfloats}
\usepackage{url}
\usepackage{verbatim}
\usepackage{graphicx}
\usepackage{cite}
\usepackage{xcolor}

\usepackage[
  separate-uncertainty = true,
  multi-part-units = repeat
]{siunitx}

\begin{document}

\title{GraphCompNet: A Position-Aware Model for Predicting and Compensating Shape Deviations in 3D Printing}

\author{
Juheon Lee, 
Lei (Rachel) Chen,
Juan Carlos Catana, 
Hui Wang, ~\IEEEmembership{IEEE member}
and Jun Zeng,~\IEEEmembership{IEEE member} 
        
\thanks{© IEEE 2025.
This is the author’s accepted manuscript of a paper published in IEEE Transactions on Automation Science and Engineering..}
\thanks{The final published version is available at: \\https://doi.org/10.1109/TASE.2025.3636394}}

\markboth{© IEEE 2025.
This is the author’s accepted manuscript of a paper published in IEEE Transactions on Automation Science and Engineering.}%
{Shell \MakeLowercase{\textit{et al.}}: A Sample Article Using IEEEtran.cls for IEEE Journals}

\IEEEpubid{0000--0000/00\$00.00~\copyright~ IEEE 2025}

\maketitle


\begin{abstract}
Shape deviation modeling and compensation in additive manufacturing (AM) are pivotal for achieving high geometric accuracy and enabling industrial-scale production. While traditional analytical and statistical methods laid the foundation, recent advancements in machine learning (ML) have improved prediction and compensation precision. However, critical challenges persist, including generalizability across complex geometries and adaptability to position-dependent variations in batch production. Traditional methods of controlling geometric deviations often rely on complex parameterized models and repetitive metrology, which can be time-consuming yet not applicable for batch production. In this paper, we present a novel, process-agnostic approach to address the challenge of ensuring geometric precision and accuracy in position-dependent AM production. The proposed GraphCompNet presents a novel computational framework integrating graph-based neural networks with a generative adversarial network (GAN)-inspired training paradigm. The framework leverages point cloud representations and dynamic graph convolutional neural networks (DGCNNs) to model intricate geometries while incorporating position-specific thermal and mechanical variations. A two-stage adversarial training process iteratively refines compensated designs using a compensator-predictor architecture, enabling real-time feedback and optimization. Experimental validation across various shapes and positions demonstrates the framework’s ability to predict deviations in freeform geometries and adapt to position-dependent batch production conditions, significantly improving compensation accuracy (35\% to 65\%) across the entire printing space, addressing position-dependent variabilities within the print chamber. The proposed method advances the development of a Digital Twin for AM, offering scalable, real-time monitoring and compensation capabilities. This work bridges critical gaps in AM process control, paving the way for high-precision, automated, and industrial-scale design and manufacturing systems.

\end{abstract}

\textit{Note to Practitioners}—This paper introduces a framework for predicting and compensating position-dependent shape deviations in 3D printing. The proposed approach effectively models complex and arbitrary 3D geometries while addressing variabilities across different printer positions, making AM more suitable for industrial batch production. This advancement demonstrates the potential to integrate Digital Twin technology into AM processes, enabling closed-loop design optimization and enhancing both precision and scalability in large-scale, practical applications. The code for this framework is available on the NVIDIA Modulus platform: https://github.com/NVIDIA/modulus.


\begin{figure}
\centering
\includegraphics[width=0.48\textwidth]{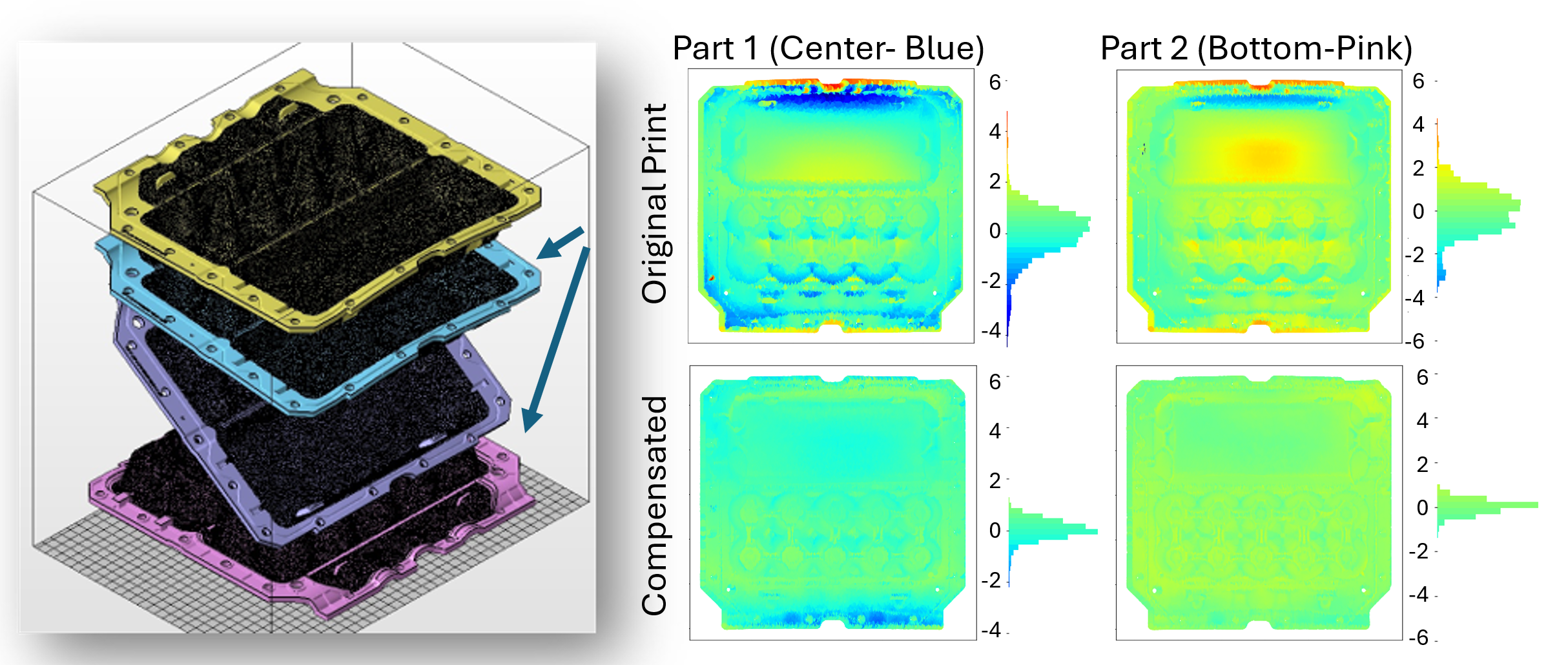}%
\caption{Configuration of the Molded Fiber dataset within the printing chamber, illustrating part placement, including stacking in the y-orientation and rotation in the x-orientation. The top row displays the deviations of two identical parts placed at different positions within the same print bucket, with a heatmap indicating the scale of deviation. These parts display distinct deviation patterns across their geometries, as highlighted by the varying color map distributions. The bottom row shows the decreasing deviations after compensation using the proposed compensation framework in both parts.}
\label{fig1}
\end{figure}

\IEEEpubidadjcol
\section{Introduction}
\IEEEPARstart{A}{dditive} 
 manufacturing (AM), commonly known as 3D printing, encompasses a diverse set of layer-by-layer fabrication processes capable of producing complex geometries with high dimensional precision. Among these, Powder Bed Fusion (PBF)—including Laser Powder Bed Fusion (L-PBF) and Multi Jet Fusion (MJF) \cite{mjf-bio}—has emerged as a leading platform due to its fine spatial resolution (typically 80–250 microns) and ability to batch-produce intricate, high-performance components. Despite these advantages, PBF and other AM processes face persistent challenges in geometric fidelity, such as thermal distortion, shape deviations, and position-dependent variability across the build platform. These issues are common across AM methods and remain critical bottlenecks for industrial-scale deployment.


For example, Fig. \ref{fig1} illustrates geometric deviations for two vertically stacked parts printed simultaneously in a multi-part build: Part 1 (Center - Blue) and Part 2 (Bottom - Pink). The heatmaps reveal distinct distortion patterns for each part, highlighting position-dependent deformation within the same build. Part 1 exhibits complex warping with edge uplift and non-uniform shrinkage, whereas Part 2 shows a more symmetric dome-like distortion with fewer localized defects. Such differences likely stem from variations in thermal boundary conditions and support interactions during printing.


Achieving high precision in AM is challenging due to the complex interactions of multiple factors, including material properties, part orientation, printer settings, calibration, layer-by-layer deposition, energy distribution, temperature gradients, and residual stresses, among others. Early research focused on analytical and statistical models to predict and mitigate these deviations, but their reliance on simplified assumptions limited applicability to real-world geometries. Recent advances in computational methods, particularly machine learning (ML), have introduced data-driven models that enhance prediction accuracy and scalability. Despite progress, critical challenges hinder the transition of these models to industrial-scale AM. 

Current deviation prediction models often struggle to generalize across complex geometries. Finite element analysis (FEA) methods \cite{10.1115/1.4041626} are computationally intensive and unsuitable for real-time applications, while many data-driven approaches have only been validated on simple shapes (e.g., domes). Moreover, most compensation models do not explicitly consider position-dependent variations (referred to here as "position-aware"), such as thermal gradients across the build plate. The lack of extensive experimental validation further limits practical deployment.

Because batch production is essential for industrial AM, addressing these challenges is critical for enabling automated systems like Digital Twins that enhance precision, control, and scalability. Such advancements would substantially improve design fidelity and production reliability, accelerating innovation in digital manufacturing.

This paper proposes a generalizable framework for modeling and compensating geometric deviations in AM, using PBF as the primary use case. The approach avoids reliance on process-specific assumptions, making it applicable to other AM platforms experiencing similar spatial and thermal effects.
The rest of the paper is organized as follows: Section II reviews related work in geometric deviation prediction and compensation, highlighting key research challenges. Section III introduces the proposed methodology. Section IV presents experimental validations and performance analysis, and Section V concludes the paper with a summary and future research directions.

\section{State-of-the-art and research challenges}

Shape deviation modeling and compensation are critical for improving geometric accuracy and quality control in AM. Early work focused on analytical and statistical models, while recent advances leverage data-driven methods such as machine learning. This section highlights three key areas: (1) Deviation Prediction Modeling, (2) Deviation Compensation Modeling, and (3) Research Challenges and Our Proposed Methodology, emphasizing the shift to data-driven strategies and the hurdles in achieving industrial-scale precision in AM mass production.

\subsection{Deviation Prediction Modeling}

Early research on shape deviation statistical and analytical parameterization models in 2D and 3D\cite{ZHU2019496, xu2017reverse,Afazov02012021}. Schmutzler et al. \cite{schmutzler2016compensating} framed deviation prediction as a non-rigid shape registration problem, using B-spline registration to predict deviations and compensate CAD parts by adjusting vertices oppositely \cite{luan2017statistical}. Huang et al. \cite{huang2014statistical} used Polar and Spherical Coordinate Systems for in-plane and out-of-plane deviation prediction, though their methods required extensive calibration, limiting scalability and adaptability. Finite element analysis (FEA) was also used to simulate deformation based on process parameters \cite{10.1115/1.4041626}.


More recently, machine learning (ML) techniques have been incorporated into deviation prediction to improve accuracy. Ferreira et al. \cite{ferreira2019automated} used Bayesian neural networks to model and compensate for deviations. Decker et al. \cite{decker2021geometric} developed a random forest model for 3D freeform shapes, enabling rapid, minimal human intervention. Convolutional neural networks (CNNs) \cite{10015471} and Shape Deviation Generators \cite{8957379} have been applied to detect complex deviation patterns, while Wang et al. \cite{9786748} extended ML to both smooth and non-smooth shapes. Li et al. \cite{LI2021101695} proposed in-situ deviation monitoring during printing.

 Despite recent advances, several challenges persist. Analytical and statistical models remain computationally intensive, with simulations for a single part often requiring hours. Although machine learning approaches demonstrate potential, they are frequently restricted to simpler geometries—such as cylinders and polyhedrons—leaving a gap in effectively handling more complex 3D shapes. This underscores the need for more efficient, generalized models that can address both computational demands and the geometric complexity encountered in practical additive manufacturing applications.


\subsection{Deviation Compensation Modeling}

Input file modification methods have emerged as an alternative to directly correct deviations at the STL file level. Techniques like the Vertex Translation Algorithm (VTA) \cite{navangul2013error} and the Surface-based Modification Algorithm (SMA) \cite{zha2015geometric} iteratively refine geometric accuracy by minimizing chordal and staircase errors. However, these methods are computationally heavy, often increasing STL file sizes and requiring several iterations to meet tolerance standards. 
Afazov et al. \cite{AFAZOV2017269} proposed a method to reduce residual stresses and distortions by interpolating 3D scan data, using a mathematical model to reverse distortions and pre-distort the CAD model. Their work demonstartes the effectiveness in reducing distortion on various parts printed with laser powder bed fusion from $\pm \SI{300}{\mu m}$ to $\SI{\pm 65}{\mu m}$ micron tolerance. In a further development \cite{AFAZOV201715}, they integrated finite element analysis (FEA) with a thermal model for selective laser melting (SLM), although the computational runtime limited its industrial applicability.

To address these challenges, advanced algorithms have been developed. Bayesian models \cite{ferreira2019automated} leverage uncertainty quantification to correct geometric errors in stereolithography printing, improving robustness across different printers. Decker et al. \cite{decker2021geometric} employed random forest models to predict deviations in fused deposition modeling (FDM) using small training datasets, achieving over 44\% shape deviation reduction, though requiring geometry-specific predictor variables. Hong et al. \cite{HONG2021101594} applied artificial neural networks (ANN) to enhance dimensional accuracy in selective laser melting of truss lattice structures, demonstrating the potential of data-driven approaches for flexible freeform geometries and improved compensation.
Despite these advances, most compensation methods focus on individual shapes, neglecting the significant thermal effects that vary with part position in the build chamber. Modeling these position-dependent deviations demands extensive resources, even for a single geometry.

\subsection{Research Challenges and Proposed Methodology} 

Despite significant progress in shape deviation modeling and compensation, major challenges remain in scalability, data dependency, and generalization across varied geometries and process conditions. Addressing these gaps is critical to improving the efficiency and accuracy of additive manufacturing (AM). This work proposes a computational framework that predicts and compensates for geometric deviations by explicitly incorporating position-specific factors, such as spatial energy variations within the build chamber, for the same geometry. Given the limited prior research on position-dependent compensation in AM, this contribution is especially timely and impactful. The key challenges in this context include:

\begin{figure}
\centering
\includegraphics[width=0.48\textwidth]{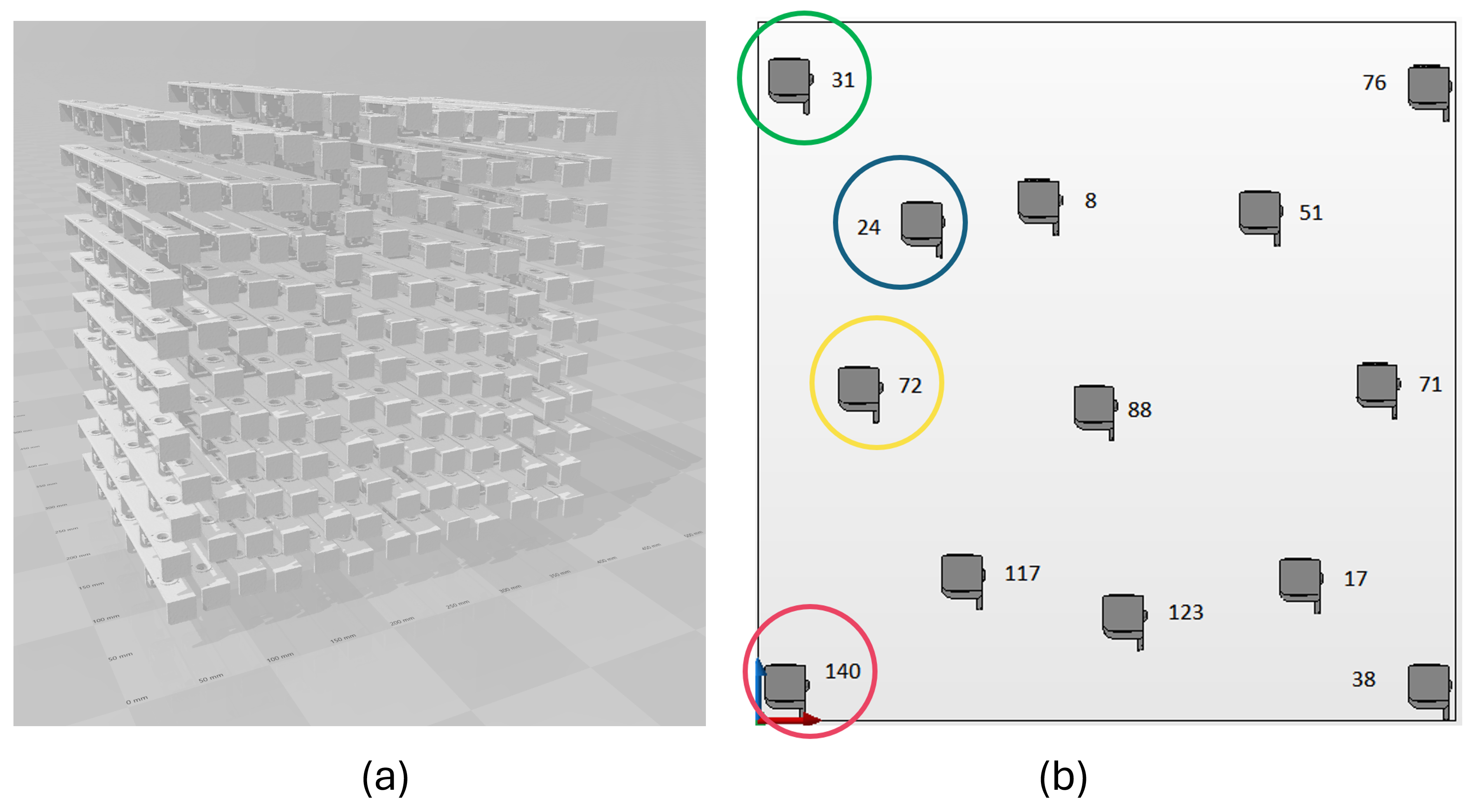}%
\caption{
The configuration of the bar dataset within the printing chamber (a) illustrates the distribution of identical part geometries placed at different positions for batch production, each labeled with a part ID. The selected training and validation parts for the case studies are highlighted with circles, as shown in (b), a cross-section of the build bed layout.}
\label{fig5}
\end{figure}

\begin{itemize}
\item \underline{\textit{Generalizability Across Complex Printing Geometries: }} 
Existing works on shape error compensation have demonstrated success in modeling, predicting, and compensating deviations for simple or structured geometries, such as 2D shapes or basic 3D forms like domes and stacked rectangles \cite{8256230, 3106707f09f04e2982a488210eaf1bdb, 8957379, 9786748}. However, these methods suffer from several key limitations that prevent generalizability to arbitrary, non-smooth 3D freeform shapes.
Finite Element Analysis (FEA)-based models are computationally expensive and impractical for complex industrial geometries and batch processes \cite{jmmp4040112}. More recent approaches, utilizing neural networks \cite{ZHU2020534, inproceedings_shen}, preprocess geometries into fixed-size images or binary formats, which limits their resolution and results in a loss of detail.

\item \underline{\textit{Adaptability to Position-Dependent Batch Production:} }
Traditional and machine learning-based compensation models often fail to consider the dynamic thermal and mechanical factors that influence AM, leading to suboptimal results under varying conditions \cite{MCGREGOR2021101924}. Thermal-induced deformations significantly impact part distortion. Research by Chen et al. \cite{10383123} shows how thermal variations across the print bed result in significant differences in part quality. For example, Figure \ref{fig5} illustrates a dataset consisting of 140 identical bars placed at different positions within the printing bed. The shape deviation patterns exhibit significant variation, with parts positioned at the edges of the print chamber, such as Part ID 140 and Part ID 31 in Fig. \ref{fig5} (b), experiencing greater deviations compared to those located at the center, such as Part ID 72 and Part ID 24. Hartmann et al. \cite{technologies7040083} highlighted the importance of batch production in industrial settings, demonstrating a data-driven compensation approach applied to stacked fin-shaped parts. However, their method is constrained by the need for iterative redesigns and was validated using only a single geometry.

This research identifies this gap in experimental validation, mainly due to high costs and resource-intensive processes. The importance of position-specific deviation modeling within the printer is often overlooked, making it a significant challenge for automation in industrial batch production. The lack of position awareness in current models is a critical barrier to implementing a Digital Twin for AM, which is necessary for improving precision and control throughout the industrial production process.
\end{itemize}

\begin{itemize}
    \item \underline{\textit{The Proposed Method}}
\end{itemize}

To address the two key challenges outlined above, we introduce a computational framework leverages \textbf{graph} data structure-based neural \textbf{net}work architecture to learn position-specific information and model shape deviations from print scan data and corresponding CAD files. The proposed \textbf{Comp}ensator-Predictor learning framework (GraphCompNet), inspired by Generative Adversarial Networks (GANs), enables the generation of compensated designs directly for any part position within the print chamber.

\begin{enumerate}

    \item To address the first challenge, to enable shape deviation prediction and compensation for arbitrary, complex geometries, we introduce two innovations:
    \begin{enumerate}
        \item \textbf{Point Cloud Data Representation}: A major challenge in modeling shape deviation and compensation for complex geometries stems from the limitations of structured data formats like images or fixed 3D structures (e.g., voxels and meshes). These formats struggle with resolution constraints for inputs of varying sizes and computational inefficiencies due to their discretized nature. To overcome these issues, we adopt point cloud data, which allows for more efficient handling of irregular and sparse data typically encountered in industrial 3D printing. Point clouds enable the model to effectively manage variations such as object orientation, scale, and differing local densities found in real-world geometries. Additionally, point clouds naturally capture spatial relationships between data points, making them well-suited for topological pattern analysis—a key element in accurately predicting shape deviations in complex geometries.
        \item \textbf{DGCNN (Dynamic Graph Convolutional Neural Network)}: To further enhance the model’s ability to handle the complexities of point cloud data, we incorporate DGCNN\cite{wang2019dynamic} as the backbone architecture.  DGCNN dynamically constructs graph structures by connecting points based on local geometric relationships, allowing it to effectively capture the varying densities and topologies inherent in complex shapes. The Edge Convolution (EdgeConv) operation—defined as a function that aggregates features from both a point and its neighbors by combining their relative and absolute positions—adapts to these local variations, facilitating effective feature extraction across the graph. This enables the model to achieve higher performance in tasks that require precise understanding of local geometry and complex spatial relationships, crucial for accurate shape prediction and deviation analysis in diverse 3D structures. 
    \end{enumerate}

    \item To address the second challenge, we incorporate a two-step approach to adapt to position-dependent factors:
    \begin{enumerate}
        \item \textbf{Incorporation of Part Position Information}: Unlike standard shape classification or segmentation tasks—where invariance to global position is often desirable—our problem formulation relies heavily on the absolute spatial location of each part within the build volume. Thermal gradients, heat dissipation patterns, and residual stress accumulation are all influenced by a part’s position on the build plate. To capture these effects, we explicitly encode part placement information in the input data. Specifically, for parts translated along the x and y axes, we adjust the corresponding point cloud coordinates to account for translational offsets due to different positions on the build plate. For parts that are rotated, we apply a rotation matrix to the point cloud coordinates. This preprocessing step preserves both the intrinsic geometry of each part and its global location within the build chamber. By embedding this positional context into the model input, we enable the network to learn spatially dependent deformation behaviors, improving predictive accuracy under varying thermal and process conditions.
        \item \textbf{Adversarial Training Framework}:Inspired by Generative Adversarial Networks (GANs), we propose a novel approach where the generation of compensation plans and the evaluation of new geometries occur simultaneously during training. This dual-process framework allows the model to effectively generate and assess results in real-time. Specifically, we introduce two networks: one learns to generate geometries that closely resemble the optimal compensated design, while the counterpart network identifies designs that will deform into the ideal outcome. This adversarial setup encourages continuous refinement of the generated geometries. 

    \end{enumerate}
\end{enumerate}

\begin{figure}
\centering
\includegraphics[width=0.48\textwidth]{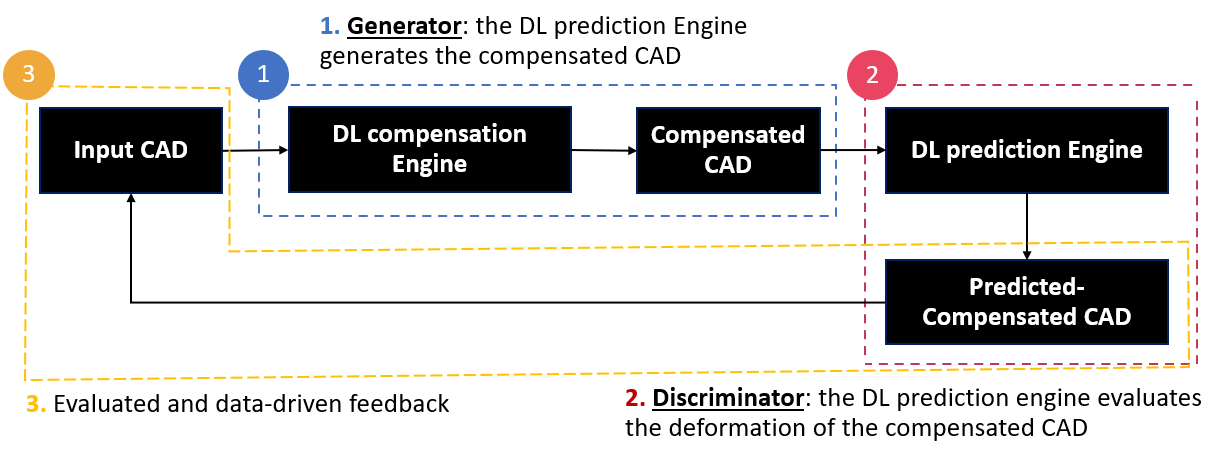}%
\caption{
The schematic diagram illustrates the proposed architecture for shape deviation prediction and compensation, inspired by the generative adversarial network framework. Initially, the DL compensation engine generates a compensation plan (step 1), which is subsequently evaluated by the DL prediction engine through shape deviation prediction (step 2). Finally, the process iterates as needed, incorporating data-driven feedback (step 3).
}
\label{fig2}
\end{figure}

Based on the findings above, this paper addresses Challenge 1 by leveraging point clouds to efficiently represent the varying densities and complex topologies inherent in industrial printing geometries, with adaptable resolution requirements. Additionally, backbone network architectures, such as DGCNN, are utilized to process point cloud data, facilitating the efficient learning of both local and global geometric relationships from the input geometries. Furthermore, this paper addresses Challenge 2 by directly learning position-based variabilities from the preprocessed input data. The GraphCompNet framework continuously proposes new geometries and assesses the proposed compensated geometries to meet the print deformation conditions. This approach overcomes the inherent limitation of machine learning models, which typically require large datasets for training, and reduces the need for extensive experimental validations. By leveraging position-aware inputs and adversarial training, the proposed framework enables closed-loop training and inference for batch production. This not only improves the efficiency and accuracy of the compensation process but also lays the foundation for implementing a Digital Twin for AM production, offering enhanced precision and control over the entire production process.




\section{Methods}
\textit{Review of Geometric Deep Learning.}
Convolutional Neural Networks (CNNs) have driven major advances in computer vision and NLP, but their application to 3D design data is limited due to the non-Euclidean nature of such structures. Graph Neural Networks (GNNs) offer a natural alternative by operating on graph-structured data $G = {\mathcal{V}, \mathcal{E}}$, where $\mathcal{V}$ and $\mathcal{E}$ represent vertices and edges, respectively. This framework allows for modeling of irregular geometries in CAD models and point clouds.

Graph neural networks can be broadly categorized into two families: spectral and spatial. Spectral methods define convolutions in the frequency domain using graph Laplacian eigen-decomposition \cite{bruna2013spectral}, though they are computationally intensive and less flexible for large-scale or dynamic graphs. Approximations using polynomial filters have been proposed to address this \cite{levie2018cayleynets}. In contrast, spatial GNNs perform convolutions through localized message passing in the node domain, making them more scalable and suited for complex, irregular graphs \cite{monti2017geometric}. A typical layer updates each node’s features by aggregating information from its neighbors:

\begin{equation}\label{eq:gnn}
    x_i^{(l)} = \Sigma_{j \in \mathcal{E}_i} h_\theta^{(l-1)} (x_i^{(l-1)},x_j^{(l-1)})  
\end{equation}
Here,  $x_i^{(l)}$ denotes the representation of node $i$ at convolutions layer $l$, $h_\theta$ is a learnable aggregation function. Each node updates its representation by aggregating features from its immediate neighbors, taking into account both node attributes and edge information $e_{ij} \in \mathcal{E}$. :

Prominent spatial GNN architectures include GraphSAGE\cite{DBLP:journals/corr/HamiltonYL17}, Graph Attention Networks (GAT)\cite{veličković2018graphattentionnetworks}, and various Graph Convolutional Network (GCN) variants. EdgeConv introduces an edge-aware operation by concatenating each node’s feature $x_i^{(l)}$ and $x_i^{l} - x_j^{(l)}$,  representing edge information with relative distance between neighboring vertices \cite{wang2019dynamic}. MoNet \cite{monti2017geometric} projects neighboring points onto a pseudo-coordinate plane, then define convolution operations on projected points on the plane. Overall, the evolution of graph neural networks presents a promising path for tackling geometric applications, offering flexibility and scalability in handling non-Euclidean data structures inherent in 3D designs \cite{Chen_2024}.

In our proposed framework, we adopt a spatial GNN due to its effectiveness in capturing localized deformation patterns and flexibility in handling varying mesh topologies. While spectral methods were considered, their computational cost and limited adaptability to local spatial variation make them less practical for our application.

\begin{figure*}[h!]
\centering
\includegraphics[width=0.8\textwidth]{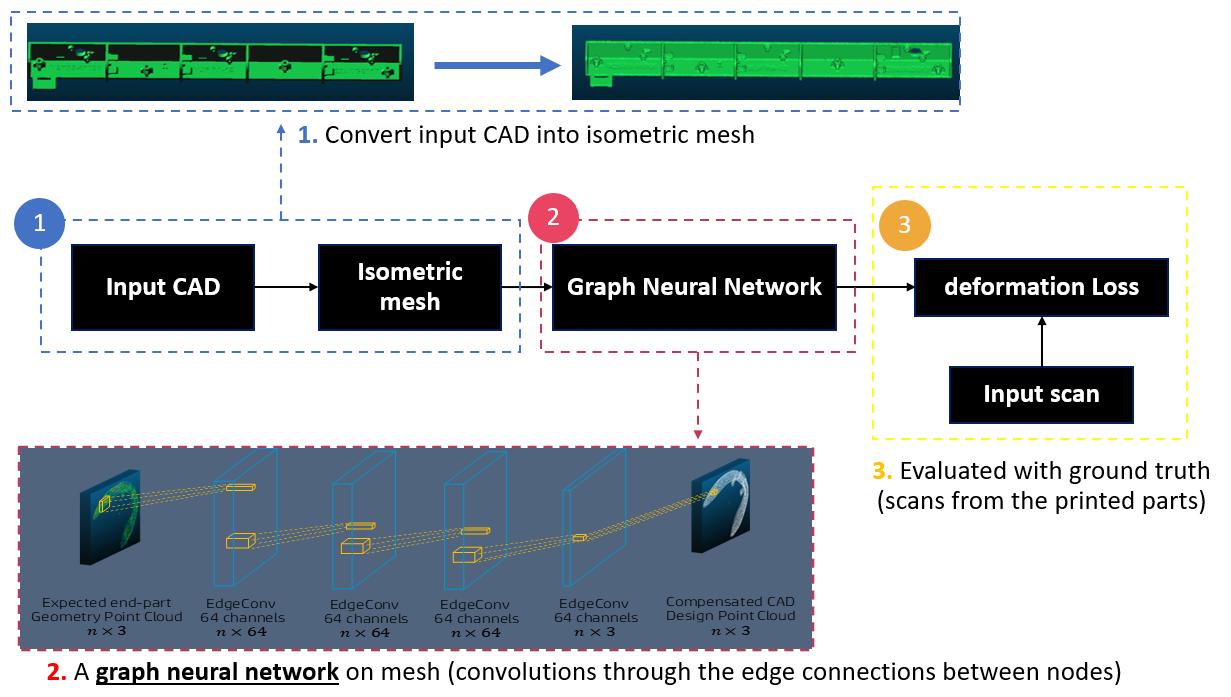}%
\caption{The procedural sequence of the DL prediction engine begins with the conversion of the input CAD into an isometric mesh ($\mathcal{C}_i = \{\mathbf{c}_1, \dots, \mathbf{c}_k\}$). Subsequently, graph neural networks are applied to predict shape deviation ($\hat{\mathcal{S}}_i = \{\hat{\mathbf{s}}_1, \dots, \hat{\mathbf{s}}_k\}$). The accuracy of the predicted shape deviation is assessed by comparing it with the printed and scanned shapes ($S_i = \{\mathbf{s}_1, \dots, \mathbf{s}_k\}$), utilizing a deformation loss function defined in Equation \eqref{eq:total_pred_loss}.}
\label{fig3}
\end{figure*}

\textit{Review of Generative adversarial network (GAN).} Generative adversarial networks (GANs) have become a potent method for modeling intricate data distributions from low-dimensional latent spaces. Recent advancements in generative deep learning have resulted in a proliferation of innovative applications across various domains, including fashion, graphic design, art, architecture, and urban planning. The GAN framework consists of two neural network components: the generator and the discriminator.

The generator maps a fixed noise distribution $\mathbb{P}_z$ to the generated data distribution $\mathbb{P}_G$, while the discriminator distinguishes samples from the distributions of generated data $\mathbb{P}_G$ and real data $\mathbb{P}_r$. GANs are trained iteratively in a minimax game, aiming to optimize the following objective function:

\begin{equation}
    \min_G \max_D \mathbb{E}_{X \sim \mathbb{P}_r}[\log D(X)]  + \mathbb{E}_{Z \sim \mathbb{P}_Z}[\log (1-D(G_\theta (Z)))]
\end{equation}
Here, $G$ and $D$ represent the generator and discriminator, respectively. While the optimization problem above is equivalent to minimizing the Jensen-Shannon divergence (JSD) \cite{goodfellow2014generative} when the discriminator is perfectly trained, JSD often leads to unstable GAN training, especially in the presence of singular measures \cite{arjovsky2017wasserstein}. To address this issue, Wasserstein-GAN was introduced, replacing JSD with the Wasserstein metric, which compares transportation costs between distributions $\mathbb{P}_G$ and $\mathbb{P}_r$.
The 1-Wasserstein distance derived from Kantorovich-Rubenstein duality is employed for GAN optimization, instead of directly computing transportation mappings. To ensure the computation of 1-Wasserstein distance, both the generator and discriminator must exhibit 1-Lipschitz continuity during training. This can be achieved through techniques such as weight clipping \cite{arjovsky2017wasserstein} or gradient penalty. 

We drew inspiration from the framework of Generative Adversarial Networks (GANs), as the tasks of prediction and compensation resemble the two components of discrimination and generation. Compensation requires accurate deviation prediction, and the compensated results must be validated through physical printing or a prediction model to assess their effectiveness.


\subsection{Data Pre-processing}
All shape deviation prediction and compensation procedures incorporate a pre-processing step involving metrology and shape registration. The printed parts undergo scanning and quality control using the ATOS scan system and GOM software \cite{VAGOVSKY20151198}. 
To enable point-level comparison with the nominal CAD geometry, we align the scanned mesh and CAD model using a combination of Deep Align
\cite{JIMENEZMORENO2021107712} and iterative closest point (ICP) algorithms \cite{article_icp}. After registration, both the CAD and scanned meshes are uniformly resampled to the same number of points, 
$k$, to standardize the input resolution for model training. This resampling ensures consistency in shape representation, though it does not assume a strict one-to-one correspondence between CAD and scan vertices.


Let $\mathbb{C}$ and $\mathbb{S}$ denote the sets of CAD models and scanned point clouds, the training dataset is comprised of pairs of CAD models and corresponding scanned point clouds, denoted as $T = {(\mathcal{C}_1, \mathcal{S}_1), \dots, (\mathcal{C}_n, \mathcal{S}_n) }$. Here, $\mathcal{C}_i \in \mathbb{C}$ represents a CAD mesh, and $\mathcal{S}_i \in \mathbb{S}$ represents the scanned point cloud derived from the printed part of $\mathcal{C}_i$. Each CAD model $\mathcal{C}_i$ is defined as a set of vertices $\mathcal{C}_i = \{\mathbf{c}_1, \dots, \mathbf{c}_k\}$, while each scanned point cloud $\mathcal{S}_i$ comprises a corresponding set of points  $\mathcal{S}_i = \{\mathbf{s}_1, \dots, \mathbf{s}_k\}$. The objective of the shape deformation prediction and compensation algorithm is to determine displacement values that predict or compensate for the shape deformation between these two sets of points. The displacement or deformation of a cloud $C_i$ is computed at the point level by pairwise matching a point from $C_i$ to a point in $S_i$, followed by the measurement of the distance between these two points using a specified metric. The simulated or predicted point cloud of the printed part with deviations is annotated as $\hat{\mathcal{S}}_i = \{\hat{\mathbf{s}}_1, \dots, \hat{\mathbf{s}}_k\}$.

The nearest neighbor algorithm is employed to compute displacement, assuming that the scanned point clouds are dense and well-aligned with the CAD models. Shape deformation is modeled through displacement mapping. Given a vertex $\mathbf{c}_i$ in the CAD part and its corresponding scanned point $\mathbf{s}_i$, the shape deformation is expressed as:
\begin{equation}
\label{eq4}
    \hat{\mathbf{s}}_i = \mathbf{c}_i + g(\mathbf{c}_i,\mathbf{\theta}) \quad \forall \mathbf{c}_i \in \mathcal{C}_i, \hat{\mathbf{s}}_i \in \hat{\mathcal{S}}_i  
\end{equation}
 where $g$ and $\mathbf{\theta}$ represent the displacement prediction function and a set of learnable parameters, respectively.

In this context, displacement mapping can be directly learned using $g$, and compensation is provided as the inverse function of $g$. However, AM processes typically exhibit non-linear behavior, where critical process variables, such as the temperature field, residual stress and melt pool geometry, spatial correlation within print part and with the neighboring geometries all impact the final outcome. This makes the above model  inadequate for proper warpage compensation and resulting in varying shape deformation outcomes \cite{schmutzler2016compensating}. To address this limitation, traditional methods often involve repeating the printing and scanning processes or employing complex model parameterization \cite{ferreira2019automated} or B-spline shape registration \cite{schmutzler2016compensating}.

\subsection{GAN-inspired Shape Deviation Compensation}

The process of shape deviation compensation and prediction mirrors the dynamics of a GAN. In the realm of shape deviation, the generator generates potential compensated shapes, while the discriminator assesses the quality of compensation based on learned shape deviation patterns. Figure \ref{fig2} illustrates the conceptual framework of our proposed architecture. Although our architecture draws inspiration from the GAN framework, it diverges from the traditional GAN concept, where the discriminator is trained to distinguish between real and generated samples.  
To avoid confusion with traditional GAN terminology, we refer to our discriminator and generator as the DL prediction and compensation engines, respectively.

In our approach, the DL prediction engine is trained to recognize shape deviation based on ground truth data (actual printed parts). Subsequently, with the parameters of the DL prediction engine fixed, the DL compensation engine is trained to propose shape compensation plans, which are then evaluated by the DL prediction engine. This iterative process continues as new or updated data becomes available, enhancing the stability of both the DL prediction and compensation engines during training. Below Eq. \ref{eq:updated_4A} indicates the predicted shape deviation for original CAD vertex, Eq. \ref{eq:updated_4B} indicates the compensated CAD vertex from the compensation engine and Eq. \ref{eq:updated_4C} indicates the predicted shape deviation for the compensated vertex.

\begin{subequations} \label{eq:updated_eq}
\begin{align}
    \hat{\mathbf{s}}_i &= D(\mathbf{c}_i; \mathbf{\theta}_{pred}) \label{eq:updated_4A} \\
    \mathbf{c}_i^{\prime} &= G(\mathbf{c}_i; \mathbf{\theta}_{comp}) \label{eq:updated_4B}\\    \hat{\mathbf{s}}_i^{\prime} &= D(\mathbf{c}_i^{\prime}; \mathbf{\theta}_{pred}) \label{eq:updated_4C}
    && 
\end{align}
\end{subequations}

\noindent\textbf{Notation Explanation:}
\begin{itemize}
    \item \(\mathbf{c}_i\): The original CAD vertex before compensation.
    \item \(\mathbf{c}_i^{\prime}\): The compensated CAD vertex proposed by the compensation engine \(G\).
    \item \(\hat{\mathbf{s}}_i\): The predicted position of the printed part surface corresponding to the original CAD vertex \(\mathbf{c}_i\) as estimated by the prediction model $D$. This does not represent a difference vector only, but rather the simulated location of the deformed surface point.
    \item \(\hat{\mathbf{s}}_i^{\prime}\): The predicted position of the printed part surface corresponding to the compensated CAD vertex \(\mathbf{c}_i^{\prime}\), also estimated by the prediction model $D$. This reflects the expected outcome after compensation and includes modeled deformation effects.
    \item \(D(\cdot; \mathbf{\theta}_{pred})\): The DL prediction engine, parameterized by \(\mathbf{\theta}_{pred}\).
    \item \(G(\cdot; \mathbf{\theta}_{comp})\): The DL compensation engine, parameterized by \(\mathbf{\theta}_{comp}\).
\end{itemize}

\noindent The learning process aims to optimize the parameters \(\mathbf{\theta}_{comp}\) of the DL compensation engine \(G\) to minimize the shape deviation between the compensated vertices \(\mathbf{c}_i^{\prime}\) and the actual scanned points \(\mathbf{s}_i\) by minimizing the deformation loss.

\begin{figure*}
\centering
\includegraphics[width=0.85\textwidth]{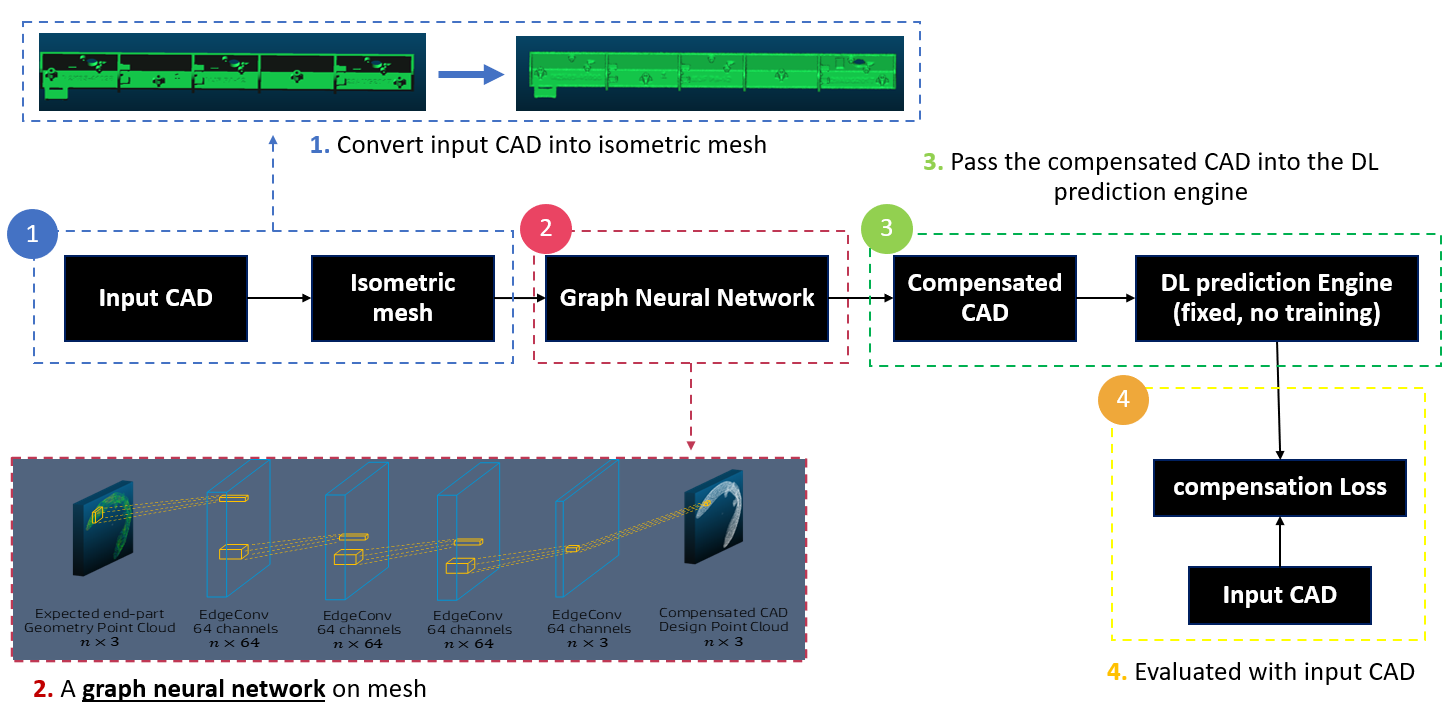}%
\caption{The workflow of the DL compensation engine. The DL compensation engine converts the input CAD into isometric mesh ($\mathcal{C}_i$) and then applies graph neural networks to find an optimal compensation plan. It passes through a weight-freeze DL prediction engine, then compares the shape deviation results with the input CAD.}
\label{fig4}
\end{figure*}


\subsection{DL prediction engine}
The DL prediction engine consists of three modules: CAD–isometric mesh conversion, graph neural network, and deformation loss computation. Fig \ref{fig3} shows the flowchart of the proposed DL prediction engine. 

\subsubsection{CAD - Isometric mesh conversion }

Geometric primitives in CAD models—such as triangles, rectangles, or hexagonal meshes—are inherently irregular, exhibiting significant variability in shape and size. While this non-uniformity offers a compact geometric description, it can negatively impact the accuracy of shape deviation prediction and compensation models. Specifically, the precision of these models is strongly correlated with the uniformity and density of the underlying geometric primitives.
To mitigate the influence of irregular primitive shapes, we convert the original CAD parts into isometric meshes using our in-house CAD-to-mesh conversion software. The resulting isometric mesh $G = \{\mathcal{V}, \mathcal{E}, \mathcal{F} \}$ is represented as a graph, where, $\mathcal{V}, \mathcal{E}, \mathcal{F}$ are sets of vertices, edges, and faces, respectively. 
Our re-meshing procedure leverages a three-dimensional discrete diffusion approximation. The input triangular mesh is first transformed into a high-density voxel grid. Surface voxels are identified by analyzing their 
3×3×3 neighborhoods. Starting from a randomly selected surface voxel, the mesh is progressively “wrapped” by expanding to adjacent voxels, including diagonal neighbors weighted probabilistically by 
$1/(4\sqrt[2]{2})$. This expansion continues until all surface voxels are encompassed, resulting in a mesh with near-uniform vertex spacing and geodesic distances.

Using isometric meshes offers several advantages compared to working with the original non-uniform meshes. The uniform vertex distribution reduces bias introduced by irregular primitive sizes and shapes, facilitating more consistent and stable graph-based learning. This uniformity leads to more stable graph-based learning and better predictive accuracy for geometric deviation and compensation. Without re-meshing, heterogeneous local geometries can hinder model performance and generalization.

\subsubsection{Graph neural network} 
Graph neural networks (GNNs) enable convolution over non-Euclidean data such as CAD surfaces represented by isometric meshes. In this work, we employ edge convolution-based GNNs \cite{wang2019dynamic}, which operate on a mesh graph $G=\left(\mathcal{V},\mathcal{E}\right)$ as follows: 
\begin{equation}\label{eq:edge_conv}
    h_\theta^{(l)} (x_i^{(l-1)}, x_j^{(l-1)}) = \frac{1}{\vert \mathcal{E}_i \vert}\sum_{j \in \mathcal{E}_i} \theta^{(l)} \cdot ( x_i^{(l-1)} \vert \vert x_i^{(l-1)} - x_j^{(l-1)})
\end{equation}
where, $(\cdot \vert \vert \cdot)$ is a concatenation operation between two tensors, $ x_j-x_i$ captures the local neighborhood information, i.e. the graph neural network provides not only global shape but also local shape information.  

Using isometric mesh as a graph provides two advantages. First of all, it is computationally cheaper than using $k$-nn graphs as it is pre-determined by existing mesh structures. Second, geometric primitives, that forms an isometric mesh, naturally encode the parts' surface geometry, such that it gives a better representation to predict shape deviation.  With the standard activation layer (e.g. ReLU) and the edge convolution $h_\theta$ defined in Equation \eqref{eq:edge_conv}, edge convolution layers are defined as:
\begin{equation}
\mathbf{x}_{i}^{\prime (l)}=\text{ReLU} (h_\theta^{(l)} (x_i^{(l-1)}, x_j^{(l-1)}))
\end{equation}

The detailed implementation of graph neural networks in the DL compensation / prediction engines are explained in the following sections.


\subsubsection{Training the DL Prediction Engine}
The DL prediction engine \(D(\mathbf{c}_i; \mathbf{\theta}_{pred})\) is trained by minimizing the deformation loss between the predicted shape deviation \(\hat{\mathbf{s}}_i\) and the actual scanned point \(\mathbf{s}_i\). The optimization problem is defined as:


\[
\min_{\mathbf{\theta}_{pred}} \quad \text{Deformation\_loss}
\]

\begin{equation}\label{eq:total_pred_loss}
\text{Deformation\_loss} = L2\_\text{Loss}  +  \text{Chamfer\_loss}
\end{equation}

The standard $L_2$ loss (a.k.a mean square loss) is used to compute regression between the isometric mesh and the deformation, represented by the scanned point clouds. Let $\mathcal{S}, \hat{\mathcal{S}}$ be the sets of vertices and corresponding point clouds from scanned part and the model predicted deviation, respectively. The standard $L_2$ loss is defined as follows:
\begin{equation}
L2\_\text{Loss}\ =\frac{1}{n} \sum_{i=1}^{n} \|\hat{\mathbf{s}}_i - \mathbf{s}_i\|^2
\end{equation}

However, $L_2$ loss does not provide shape consistency, such that it may cause some oscillations or other irregular patterns. In order to preserve shape consistency, we introduce the Chamfer loss function. Chamfer distance computes the sum of the smallest distances between each element in $\hat{\mathcal{S}}$ and $\mathcal{S}$, thus penalizing shape inconsistency between $\hat{\mathcal{S}}$ and $\mathcal{S}$. It is defined as:
\begin{equation}
\text{Chamfer\_loss} = \sum_{\hat{\mathbf{s}}_i \in \hat{\mathcal{S}}} \min_{\mathbf{s}_j \in \mathcal{S}} \|\hat{\mathbf{s}}_i - \mathbf{s}_j\| + \sum_{\mathbf{s}_j \in \mathcal{S}} \min_{\hat{\mathbf{s}}_i \in \hat{\mathcal{S}}} \|\mathbf{s}_j - \hat{\mathbf{s}}_i\|
\end{equation}

Here:
\begin{itemize}
    \item \(\mathbf{c}_i\): Original CAD vertex.
    \item \(\mathbf{s}_i\): Actual scanned point (ground truth).
    \item \(\hat{\mathbf{s}}_i = D(\mathbf{c}_i; \mathbf{\theta}_{pred})\): The predicted position of the printed part surface corresponding to the original CAD vertex \(\mathbf{c}_i\), as estimated by the prediction model $D$. This includes anticipated shape deviations due to process-induced distortion.
    \item \(\mathbf{\theta}_{pred}\): Parameters of the DL prediction engine to be optimized during training.
\end{itemize}

\subsection{DL compensation Engine}

\begin{figure*}
\centering
\includegraphics[width=0.8\textwidth]{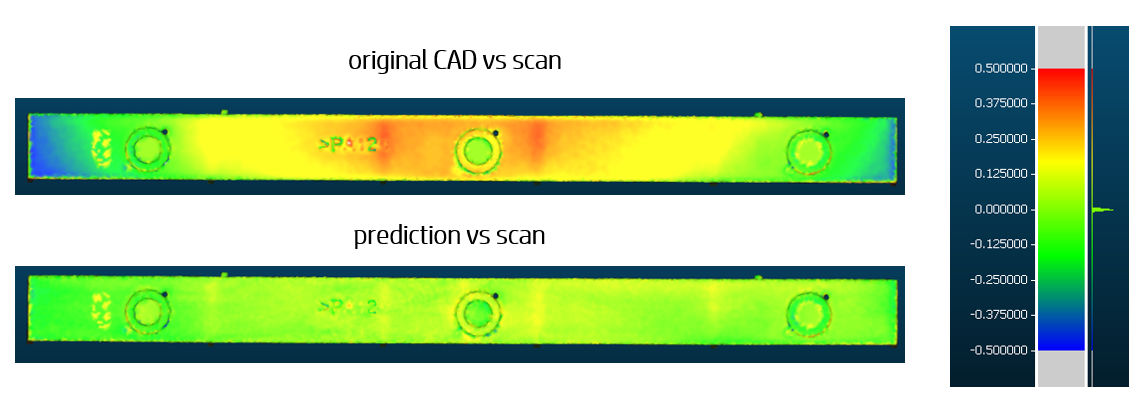}%
\caption{\textit{Prediction performance on the sample test bar, visualized from a top-down (planar) view.} The heatmaps (units in mm) represent geometric deviation magnitudes across the build plate. The \textbf{top heatmap} shows the deviation between the original CAD model and the scanned printed part, capturing actual deformation. The \textbf{bottom heatmap} compares the trained prediction engine’s output to the scanned print, illustrating the model’s accuracy. The non-uniform distortion observed in the physical print is well approximated by the prediction engine, demonstrating its ability to capture position-dependent deformation behavior.}
\label{fig6}
\end{figure*}

The DL compensation engine is designed to correct geometric distortions that occur during the printing process by adjusting the input CAD model accordingly. The adjusted CAD model produced by this engine is then used for printing. During the training phase, the compensated CAD model is passed to the well-trained DL prediction engine, which estimates the shape deviation of the adjusted model. Successful compensation is indicated by a minimized shape deviation between the compensated model and the original input CAD model, as predicted by the DL engine.

Comprising four principal modules, the DL compensation engine is structured as follows: 1. CAD-to-isometric mesh conversion, 2. Graph neural network,
3. DL prediction engine, and
4. Evaluation.
The design of the DL compensation engine as illustrated in fig \ref{fig4} shows the flowchart of the proposed DL prediction engine. mirrors that of the DL prediction engine. The procedure for CAD-to-isometric mesh conversion aligns with the methodology described in the DL prediction engine section. Central to the DL compensation engine is the graph neural network, which is adapted from the architecture utilized in the DL prediction engine. Minor adjustments have been implemented to enhance model performance. Evaluation of the DL compensation engine's accuracy is conducted by computing the deformation loss between the compensated-predicted shape deviation and the input CAD, as follows:



\begin{equation}
L2\_\text{Loss} (\mathcal{C}_1, \hat{\mathcal{S}}_1^{\prime})\ =\ \frac{1}{n}\sum_{i=1}^{n}{||\mathbf{c}_i - \hat{\mathbf{s}}_i^{\prime}||^2}
\end{equation}

 {\small
\begin{align}
 &\text{Chamfer\_loss} (\mathcal{C}_1,\hat{\mathcal{S}}_1^{\prime}) = \nonumber\\
 &\sum_{\mathbf{c}_i\in \mathcal{C}} \min_{\hat{\mathbf{s}}_j^{\prime} \in \hat{\mathcal{S}}^{\prime}} \Vert \mathbf{c}_i - \hat{\mathbf{s}}_j^{\prime} \Vert 
 + \sum_{\hat{\mathbf{s}}_j^{\prime} \in \hat{\mathcal{S}}^{\prime}} \min_{\mathbf{c}_i \in \mathcal{C}} \Vert \hat{\mathbf{s}}_j^{\prime}  - \mathbf{c}_i \Vert
\end{align}}

\[
\text{Deformation loss} = L2\_\text{Loss}  +  \text{Chamfer\_loss}
\]




Here:
\begin{itemize}
    \item \(\mathbf{\theta}_{comp}\): Parameters of the DL compensation engine to be optimized during training.
    \item \(\mathbf{\theta}_{pred}\): Parameters of the DL prediction engine, which is fixed during the DL compensation engine training.
\end{itemize}

\section{Case Studies}
As noted in Section II, prior approaches struggle with complex topologies and are limited in closed-loop batch production due to high experimental costs and lack of position-specific modeling. To demonstrate the effectiveness of our Predictor-Compensator framework, we conducted case studies using real parts printed on an HP Multi Jet Fusion (MJF) system with PA12, a thermoplastic known for strength and dimensional stability. This setup reflects a common industrial environment, underscoring the framework’s practical relevance.

As shown in Figures \ref{fig1} and \ref{fig10}(a), large parts often exhibit warping from thermal gradients across the build chamber. To evaluate the effect of nesting (the strategic arrangement of objects within a 3D printer's build chamber) on distortion, Case Study A involved a nested bucket containing 140 bar-shaped parts printed at different positions. Case Study B extended the framework to a complex geometry (Molded Fiber), demonstrating adaptability to varying part orientations and shapes.
Printed parts were scanned and reconstructed using the GOM Suite (GOM, Germany). Both the deviation prediction and compensation modules use the same graph neural network architecture with four edge convolution layers (see Figures \ref{fig3} and \ref{fig4}), implemented in PyTorch and PyTorch Geometric. Models were trained using the Adam optimizer (learning rate = 0.001) for 1000 epochs on a single NVIDIA RTX 3090 GPU (24GB VRAM).

\begin{figure*}
\centering
\includegraphics[width=0.85\textwidth]{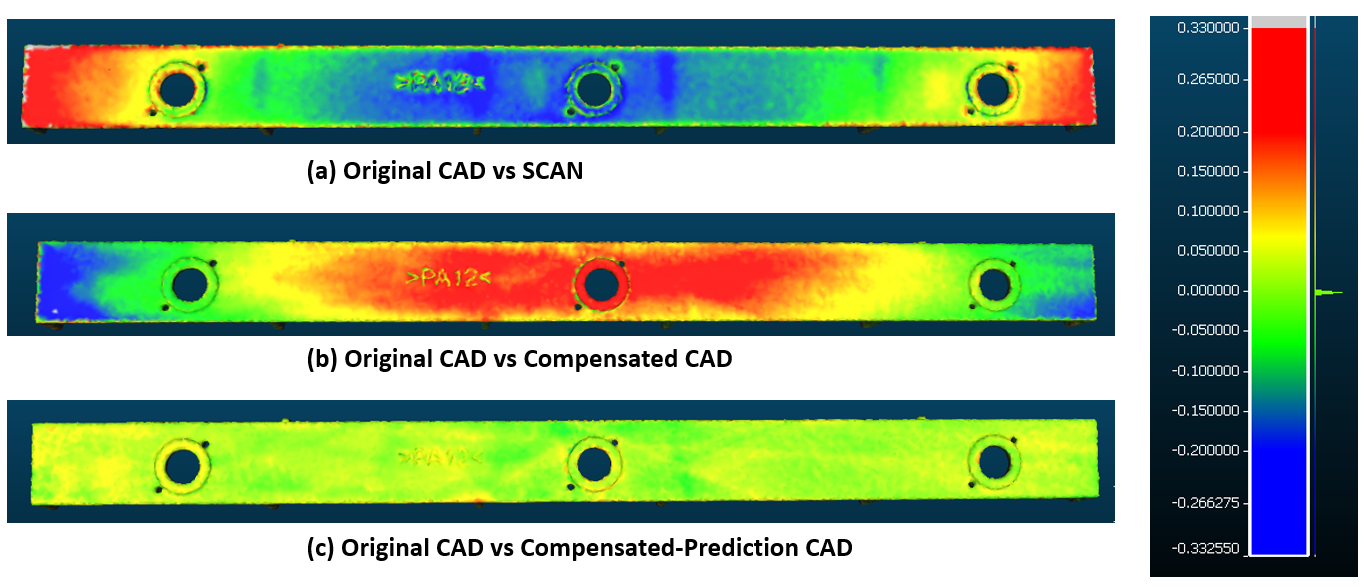}%
\caption{\textit{Performance of the proposed deep learning–based compensation engine on test data.} The heatmaps (units in mm) visualize geometric deviation magnitudes between different 3D shape comparisons. Specifically, the plots include deviations between the nominal CAD model and the scanned printed part, the nominal CAD and the compensated CAD geometry, the nominal CAD and the predicted position of the print part surface for the model proposed compensated CAD geometry. These visualizations illustrate how well the compensation strategy reduces geometric errors relative to the original design.}
\label{fig7}
\end{figure*}


\subsection{Case A: Nesting-Dependent Bar Part Warpage}

As shown in Fig.\ref{fig5}(a), 140 identical bar-shaped parts were arranged across the print bed to analyze the impact of spatial variation on shape deviations. Due to uneven thermal gradients, bars near the edges (e.g., Part IDs 140 and 31) exhibited greater deformation than centrally located parts (e.g., Part IDs 72 and 24), as illustrated in Fig.\ref{fig5}(b). Fig.\ref{fig6} provides an example of the observed warpage in ``Original CAD versus Scan", where the color gradient represents the geometric deviation between the original CAD model and the scanned printed part. In this heatmap, blue indicates negative prediction deviation, while red represents positive prediction deviation. It is evident that the center of the bar part experiences higher positive warping, whereas the edges exhibit warping in the opposite direction.

To train the prediction engine, 13 bars were randomly selected, while the remaining 127 were reserved for testing. A point cloud encoding each part’s position in the build chamber was used as input to the deep learning (DL) model. Fig.\ref{fig6}(b) demonstrates that the trained prediction engine effectively captures part warpage without prior geometric analysis. 
The predicted part warpage closely matches the scanned printed part, demonstrating that the DL prediction engine effectively minimizes the geometric distance between the predicted and scanned geometries. The DL prediction engine also demonstrates the ability to detect position-dependent shape deviations. For instance, it accurately identifies concave deformation in parts located in the upper region of the print chamber (represented by Part 24 and Part 31 in Fig. \ref{fig5}), and convex deformation in parts situated in the lower region of the printing bed (represented by Part 117 and Part 140).

\begin{figure}
\centering
\includegraphics[width=0.48\textwidth]{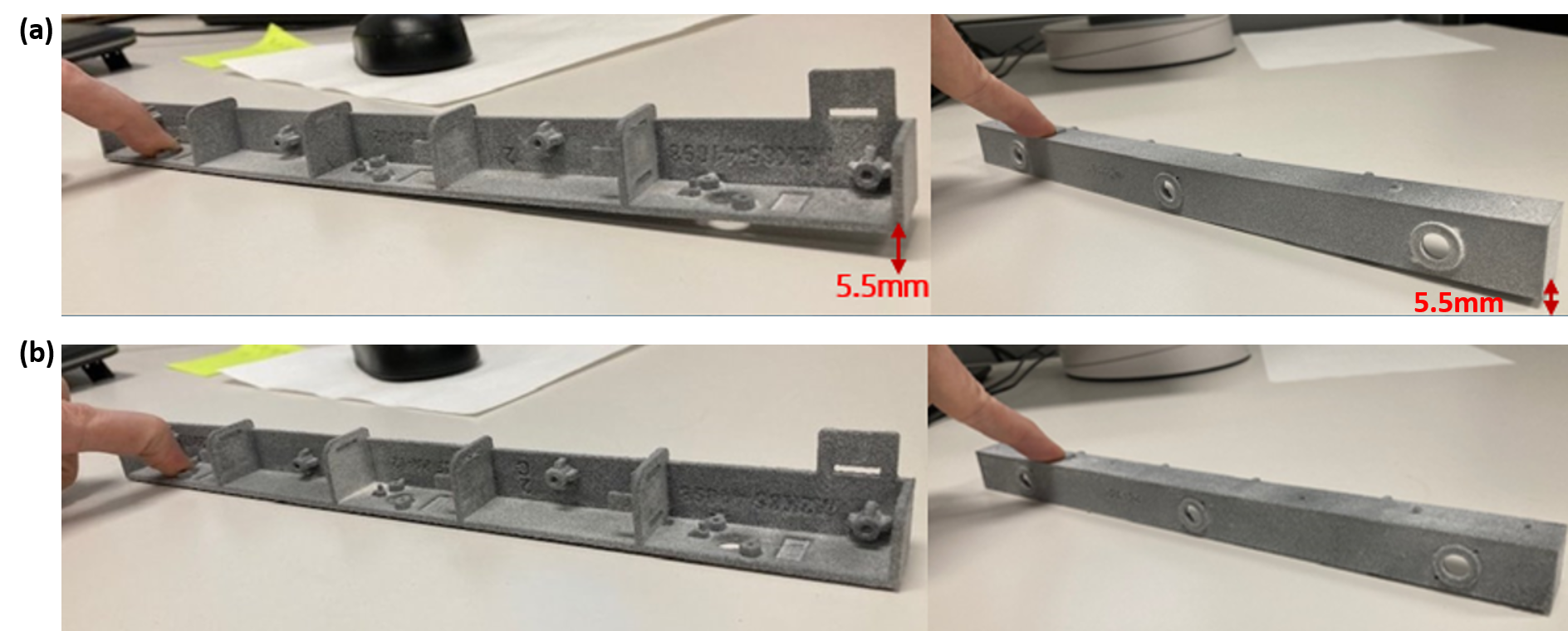}%
\caption{Comparison of the printed Bar with/without deformation compensation: (a) original CAD after print (b) compensated CAD after print. While the uncompensated part exhibits a 5.5mm warpage at the corner, the compensated part shows no noticeable warpage effect. }
\label{fig10}
\end{figure}

We next evaluated the DL compensation engine. As shown in Fig.\ref{fig7}, the original printed parts exhibit substantial deviations, especially at the corners. In contrast, the compensated CAD shows inverse adjustments, and when passed through the prediction engine (Fig.\ref{fig7}(c)), the predicted deviation is significantly reduced, closely aligning with the original CAD.


The results presented above confirm the theoretical efficacy of the proposed algorithm. However, without physical printing, there is no guarantee that the DL compensation engine has successfully corrected the CAD model in a real-world scenario. Therefore, physical validation through actual printing is essential. 
To validate physical performance, two sets of parts—compensated and uncompensated—were printed in the same batch. Post-print scanning and comparison reveal that while uncompensated parts showed up to 5.5mm warpage, the compensated parts exhibited minimal or no visible distortion (Fig.\ref{fig10}), confirming the real-world efficacy of the proposed method.


\subsection{Case B. Molded Fiber Dataset}


The molded fiber dataset consists of identical geometries designed to function as "10-egg plates." We collected multiple buckets, each containing stacked print parts with several 10-egg plates, and randomly divided these parts into training and validation sets. The sample bucket arrangements are shown in Figure 10. The "STACK" bucket contains a print arrangement with five parts stacked on top of each other. The "VERTICAL" bucket features five parts printed vertically and positioned adjacently along the y-axis. The "ROT" bucket is a modified version of the "STACK" bucket, with one randomly selected part rotated at a specific angle from the x-orientation. By varying the position and orientation of the identical geometry in the molded fiber data, we aim to demonstrate the effectiveness of our proposed model across different print runs, print locations, and print orientations. We trained the prediction and compensation engines using data from seven buckets and validated the results on parts from five buckets not included in the training set.

Figure 11 presents a visual comparison of parts from a sample validation molded fiber bucket (not used in training) containing four parts (left). The highlighted part was rotated 17 degrees downward along the X-axis. The other parts in the bucket are placed at different positions and orientations: Part 2 is translated along the X-axis, Part 3 is rotated -10 degrees along the X-axis and 7 degrees along the Y-axis, and Part 4 is rotated -4 degrees along the X-axis and -85 degrees along the Y-axis. The top right row shows the difference between the design CAD file and the scanned printed part geometry before compensation for the four parts placed in the print chamber. The bottom row illustrates the results after applying compensation using the trained GraphCompNet, highlighting the difference between the compensated CAD file and the scanned printed part geometry. The color bars are scaled identically, representing voxel-wise differences. The compensated part shows significant improvement, particularly at the top and bottom edges, where relatively large deformations were initially observed.

The quantitative improvement after applying the trained GraphCompNet to this bucket is presented in Table 1 (in mm). The table provides a detailed comparison of geometric deviations before and after compensation, showcasing the effectiveness of the trained engine in correcting shape deformations. The absolute mean deviation of the compensated parts after printing shows improvements ranging from 30\% to 66\%, with reductions in both the maximum deviation and the standard deviation, compared to parts that did not undergo compensation.

\begin{figure}
\includegraphics[width=0.48\textwidth]{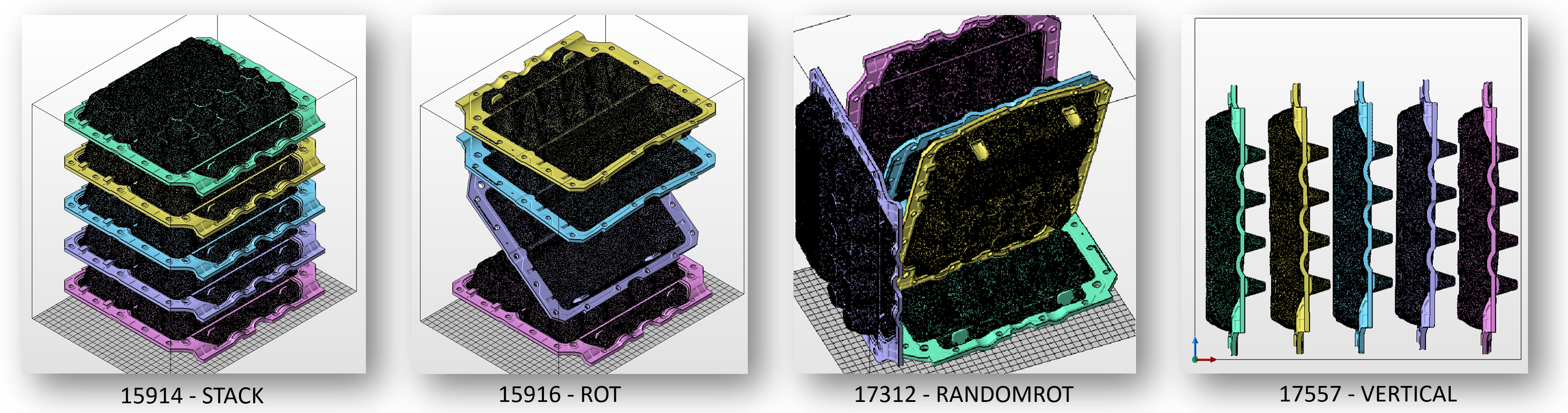}%
\caption{Sample bucket arrangement of molded fiber dataset. The "STACK" bucket contains a print arrangement with five parts stacked on top of each other. The "VERTICAL" bucket features five parts printed vertically and positioned adjacently along the y-axis. The "ROT" bucket is a modified version of the "STACK" bucket, with one randomly selected part rotated at a specific angle from the x-orientation. }
\label{fig11}
\end{figure}

\begin{table}[!t]
\caption{Quantitative improvement before \& after applying compensation \label{tab:table1}}
\centering
\begin{tabular}{|c||c|c|c|c|c|}
\hline
Metric (mm)  & Min & Max  & Std & Abs Mean & Improve  \\
\hline
Part 1 & -3.57 & 3.83  & 0.88 & 0.67 & \\
P1 compensated &\bf{-2.70} & \bf{2.45} & \bf{0.52} &\bf{0.40}& 29.9\%\\
\hline
Part 2  & -3.59 & 3.89 & 0.97 & 0.76 & \\
P2 compensated  & \bf{-1.58} & \bf{1.26} & \bf{0.33}  & \bf{0.26} & 65.8\%\\
\hline
Part 3  & -3.66 & 4.01  & 0.88& 0.65 & \\
P3 compensated & \bf{-1.71} & \bf{1.77} & \bf{0.35} & \bf{0.27} & 58.5\%\\

\hline
Part 4  & -3.40 & 3.65  & 0.80 & 0.57 & \\
P4 compensated & \bf{-1.71} & \bf{2.41} & \bf{0.32} & \bf{0.25} & 56.1\%\\
\hline
\end{tabular}
\end{table}

 \begin{figure*}
\centering
\includegraphics[width=\textwidth]{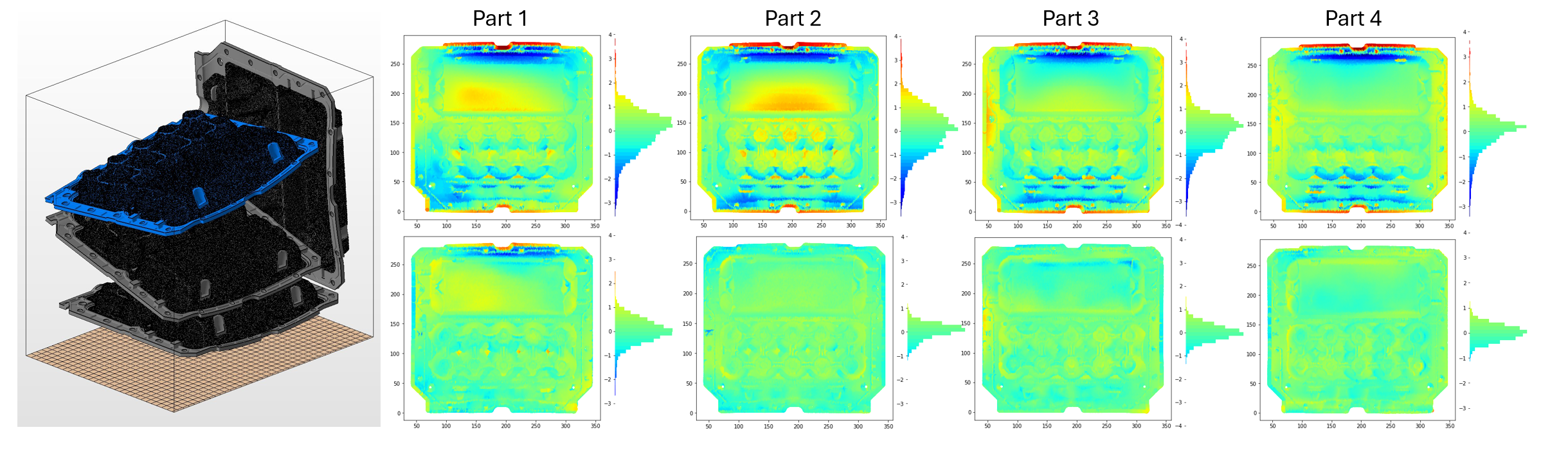}
\caption{Comparison of four sample parts in one print run, the top row illustrates the difference between the design CAD file and the scanned printed part geometry before applying compensation, the bottom row shows the difference between the design CAD file and the scanned printed part geometry after applying compensation using our trained prediction and compensation engine.}
\label{fig13}
\end{figure*}

\subsection{Comparison with State-of-the-art}

Compared to our results in Table I, the study by Decker et al. \cite{decker2021geometric} reports an average vertex error reduction of about 44\% on a single test part, which is broadly comparable to the performance of our model. However, it is important to note that this comparison is not intended as a direct benchmark, as the two studies involve different printing processes (FDM vs. PBF), which may influence deformation behavior. Instead, this reference is included to qualitatively illustrate the broader challenge of geometric compensation in additive manufacturing. In our study, three out of four parts demonstrate improved compensation performance (56\% to 66\% error reduction), with our model offering further advantages in robustness and accuracy across varying build positions and orientations. By explicitly accounting for position-dependent variables that lead to diverse deformation patterns, our approach shows greater generalizability and effectiveness under a wider range of conditions. We have revised the manuscript to emphasize that such cross-process comparisons should be interpreted with caution.

\section{Conclusions}

This paper addresses the critical challenge of achieving high geometric precision in additive manufacturing (AM), focusing on modeling and compensating for shape deviations driven by thermal gradients, residual stress, and other process-induced distortions. Existing models often lack generalization across complex geometries and fail to account for position-dependent variability in batch production.

To address these limitations, we proposed GraphCompNet, a graph-based framework that integrates (1) a DGCNN backbone to process point cloud data and capture both local and global geometric patterns, and (2) position-specific inputs within an adversarial training setup to enable closed-loop, position-aware compensation. Experimental validation across various geometries showed consistent improvement, with deviation reductions ranging from 30\% to 66\%, demonstrating strong generalization and competitive or superior performance to existing methods.

The proposed method offers a scalable and efficient alternative to traditional simulation-heavy workflows by reducing the need for geometry-specific retraining or costly experimentation. Its ability to adapt to spatial variations within the print bed further enhances applicability in industrial settings.

Future work includes:
Exploring alternative coordinate encodings (e.g., sinusoidal, Fourier) to improve model accuracy and generalization.
Evaluating the number of calibration builds needed for adapting to new printers, process parameters, or geometries.
Investigating the latency of the full compensation pipeline for deployment in real-time or near-real-time manufacturing.
Investigating adaptive or learnable loss weighting schemes to better balance local and global alignment across diverse shapes and point cloud densities.
Importantly, we emphasize the need for future studies to benchmark compensation strategies under consistent hardware, materials, and print settings. Standardized evaluation protocols will be crucial to rigorously compare methods and advance the state-of-the-art in AM deformation compensation.

In summary, this paper addresses the challenge of compensating for shape deviations in complex geometries and location-dependent variations across batch production in additive manufacturing. By leveraging deep learning techniques, specifically graph-based neural networks, our approach models both global and local geometric interactions while accounting for part-specific positioning on the printing bed. This novel compensation framework overcomes the resolution and scalability limitations of traditional methods, offering an effective solution to enhance geometric precision and production efficiency. Our method demonstrates significant potential for practical implementation in large-scale industrial production, advancing precision-driven innovation in AM processes.

\IEEEpubidadjcol

\section*{Acknowledgments}
This material is based upon work supported by the Air
Force Office of Scientific Research under award number
FA9550-23-1-0739. Any opinions, findings, and conclusions
or recommendations expressed in this material are those of
the author(s) and do not necessarily reflect the views of the
United States Air Force.


\bibliographystyle{ieeetr}
\bibliography{references}











\newpage

\section{Biography Section}
 


\begin{IEEEbiography}[{\includegraphics[width=1.0in,height=1.2 in,clip,keepaspectratio]{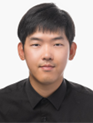}}]
{Juheon Lee} received his PhD degree from the University of Cambridge, UK, in 2016. Since 2019, he has been a research scientist at HP Inc. His research interests are in geometric deep learning physics-informed neural networks and neural tangent kernel theory.  
(juheon.lee.626@gmail.com). 

\end{IEEEbiography}

\begin{IEEEbiography}[{\includegraphics[width=1.0in,height=1.2 in,clip,keepaspectratio]{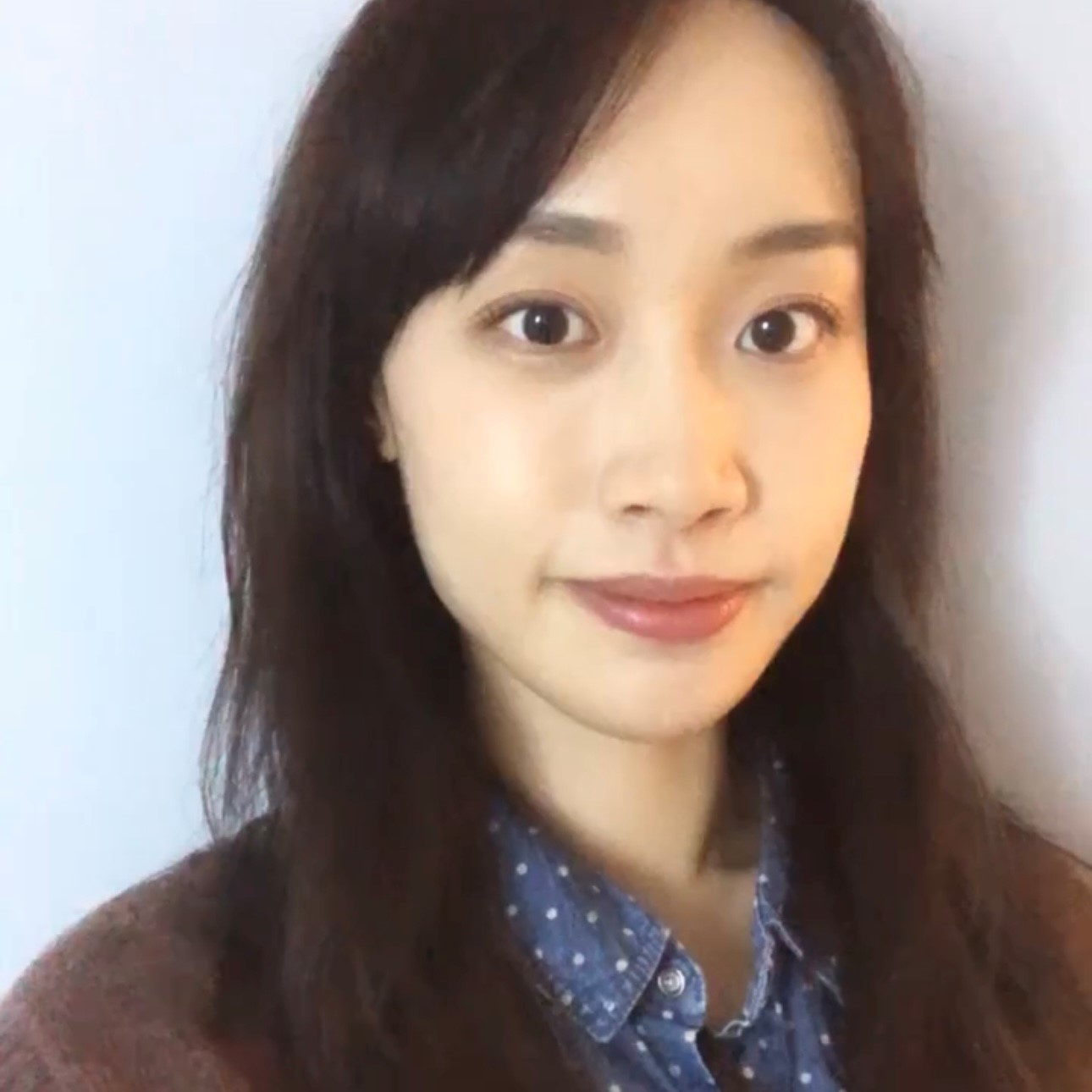}}]
{Rachel (Lei) Chen} is a Machine Learning Research Engineer at HP Inc. She received her M.S. degree in Electrical and Computer Engineering from Duke University in 2019 and her B.S. degree in Electrical Engineering from Korea Advanced Institute of Science and Technology (KAIST) in 2017. She is now devoted to research in applying cutting-edge deep learning algorithms in Additive Manufacturing quality control, acceleration, and Digital Twins. 
 (lei.rachel.chen@gmail.com)
\end{IEEEbiography}

\begin{IEEEbiography}[{\includegraphics[width=1.0in,height=1.2 in,clip,keepaspectratio]{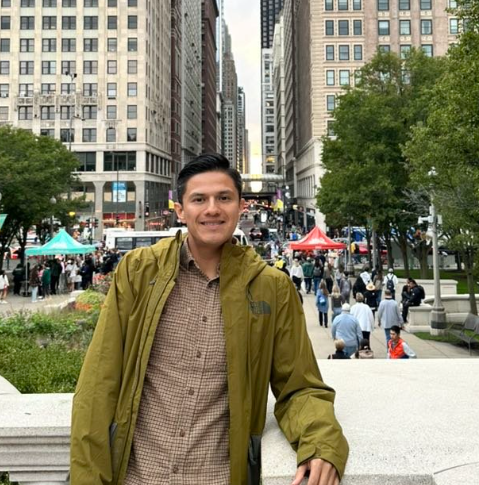}}]
{Juan Carlos Catana}, born in 1987 in Puebla, Mexico, is currently working as a Senior ML Engineer on a 3D software team at HP Inc. His primary research interests include graph algorithms, computational geometry, geometric deep learning, and parallel/distributed algorithms.
Outside of work, Juan Carlos enjoys running on the streets, trekking in the mountains, and practicing mountain biking (XC). He is passionate about traveling and discovering new places.
\end{IEEEbiography}



\begin{IEEEbiography}[{\includegraphics[width=1.0in,height=1.2 in,clip,keepaspectratio]{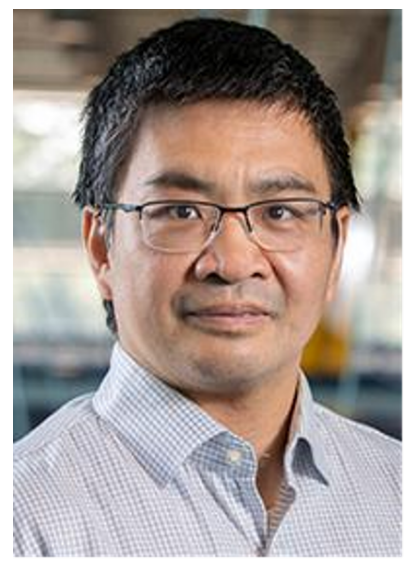}}]
{Hui Wang} is an associate professor of industrial
engineering at the Florida A\&M University-Florida
State University College of Engineering and a member of the High-Performance Materials Institute
(HPMI). His research has been focused on (i) data
modeling and analytics to support quality control
for manufacturing processes, including small-sample
learning under an interconnected environment and
optimal knowledge organization for zero-shot learning, and (ii) optimization of manufacturing system
design and supply chain. He received his PhD in
industrial engineering from the University of South Florida and an MSE in
mechanical engineering from the University of Michigan
\end{IEEEbiography}

\begin{IEEEbiography}[{\includegraphics[width=1.0in,height=1.2 in,clip,keepaspectratio]{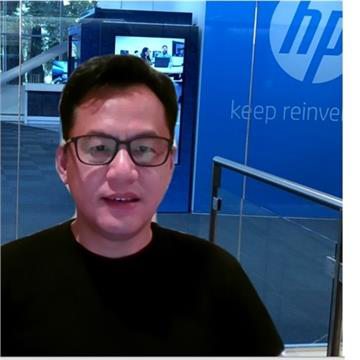}}]
{Jun Zeng} is a Distinguished Technologist at HP Inc. and a principal investigator and research manager leading software research in 3D Printing and Digital Manufacturing for the HP 3D Printing Software group. His publications include a co-edited book on the Computer-aided Design of microfluidics and biochips, a co-authored book on production management of digital printing factories, 50+ peer-reviewed papers, and 42 granted patents and more pending. His academic training includes advanced degrees in mechanical engineering (PhD) and computer science (M.S.), both from Johns Hopkins University. Jun is an ACM member and an IEEE senior member.
(jun.zeng@hp.com).
\end{IEEEbiography}

\vspace{11pt}


\vfill

\end{document}